\PassOptionsToPackage{square,comma,numbers,sort&compress}{natbib}  
\documentclass{article}

\usepackage[final]{neurips_2021}

\pdfoutput=1
\usepackage[utf8]{inputenc} % allow utf-8 input
\usepackage[T1]{fontenc}    % use 8-bit T1 fonts
\usepackage{hyperref}       % hyperlinks
\usepackage{url}            % simple URL typesetting
\usepackage{booktabs}       % professional-quality tables
\usepackage{amsfonts}       % blackboard math symbols
\usepackage{nicefrac}       % compact symbols for 1/2, etc.
\usepackage{microtype}      % microtypography
\usepackage{xcolor}         % colors
\usepackage{array}
\usepackage{arydshln}
\usepackage{bbm}
\usepackage{graphicx}
\usepackage{amssymb}
\usepackage{algorithm,algcompatible,amsmath}
\usepackage{multirow}
\usepackage{comment}

\definecolor{mygray}{gray}{0.6}

\algnewcommand\INPUT{\item[\textbf{Input:}]}%
\algnewcommand\OUTPUT{\item[\textbf{Output:}]}%
\DeclareMathOperator*{\argmax}{arg\,max}%

\title{Con$^{2}$DA: Simplifying Semi-supervised Domain Adaptation by Learning Consistent and Contrastive Feature Representations}

\author{%
  Manuel Pérez-Carrasco\\
  Department of Computer Science\\
  University of Concepcion\\
   \And
   Pavlos Protopapas\\
   IACS \\
  Harvard University \\
  \AND
  Guillermo Cabrera-Vives \\
  Department of Computer Science\\
  University of Concepcion\\
}

\begin{document}

\maketitle

\begin{abstract} 
In this work, we present Con$^{2}$DA, a simple framework that extends recent advances in semi-supervised learning to the semi-supervised domain adaptation (SSDA) problem. Our framework generates pairs of associated samples by performing stochastic data transformations to a given input. Associated data pairs are mapped to a feature representation space using a feature extractor. We use different loss functions to enforce consistency between the feature representations of associated data pairs of samples. We show that these learned representations are useful to deal with differences in data distributions in the domain adaptation problem. We performed experiments to study the main components of our model and we show that (i) learning of the consistent and contrastive feature representations is crucial to extract good discriminative features across different domains, and  ii) our model benefits from the use of strong augmentation policies. With these findings, our method achieves state-of-the-art performances in three benchmark datasets for SSDA.
\end{abstract}

\section{Introduction}

Even though deep neural networks have become the state of the art for image classification \cite{krizhevsky_2012, simonyan_2015, he_2016}, they often require many labeled data to achieve good results. Furthermore, these models usually perform poorly when test data are drawn from a different distribution or feature space. This difference between data distributions is usually called \textit{domain shift} \cite{saenko_2010, sun_2016}. To alleviate such labeling efforts, domain adaptation (DA;\cite{ben-david_2007}) aims at reducing the domain shift when a model trained on a source domain is tested on a target set sampled from a different distribution, by finding a common representation between them \cite{pan_2010}.

Semi-supervised domain adaptation (SSDA) methods address the problem by considering a labeled source and both (a few) labeled and unlabeled target data during training \cite{saito_2019, Jiang_2020, kim_2020}. For instance, Minimax Entropy (MME) \cite{saito_2019} uses a $\ell _{2}$ normalization on the output of a feature extractor and minimizes the cosine similarity between the normalized feature representation of the labeled data and a linear classifier. Thus, the weights of the linear classifier can be seen as representative points of each class (i.e. prototypes). In this work we adopted the normalized feature representation, which has proven to be effective to learn feature representations under labels restrictions \cite{gidaris_2018, chen_2019, chen_2020}.

Advances in semi-supervised learning such as consistency regularization methods \cite{laine_2017, xie_2019, berthelot_2020, chen_2020, sohn_2020} have demonstrated to improve the generalization of the model by generating the same model predictions for different randomly perturbed inputs. Also, recent works suggest that the use of heavily distorted samples (e.g., Cutout \cite{devries_2017}, or RandAugment \cite{cubuk_2020}) are effective to improve the model's performance and generalization \cite{sohn_2020}. On the other hand, recent contrastive learning approaches have shown state-of-the-art results by learning discriminative feature representations from unlabeled data. The general idea is to align a pair of positive samples and push negative samples far apart \cite{dosovi_2014, wu_2018, chen_2020, khosla_2020}.  Despite the promising results of the aforementioned approaches to learning in presence of domain shift, to the best of our knowledge, they have not been extended yet to a SSDA scenario.

In this work we adopted the normalized feature representation proposed by \cite{saito_2019}, and we extent recent advances on \textbf{Con}sistency regularization and \textbf{Con}trastive learning for the SS\textbf{DA} problem. We present \textbf{Con$^{2}$DA}, a simple framework that takes input images and generates pairs of associated augmented versions of the input. These pairs are mapped to the unit hypersphere using a feature extractor. When labels are available, we move the associated normalized representations of the pairs that come from the same class towards their correspondent class prototype. When labels are not available, we treat each pair of associated versions of the input as positive samples and pull them together, while the remaining dissimilar data pairs that come from different input images are pushed apart. An overview of our model can be seen in Figure \ref{fig:model}.

Using this simple yet effective method, we achieved state-of-the-art performances in the commonly used benchmark datasets for SSDA DomainNet \cite{peng_2019}, Office-Home \cite{office-home}, and Office-31 \cite{saenko_2010}, improving up to $1.3\%$, $1.3\%$, and $1.4\%$ in average for each dataset respectively.

\section{Method}
\begin{figure*}[t]
\centering
  \includegraphics[width=0.95\linewidth]{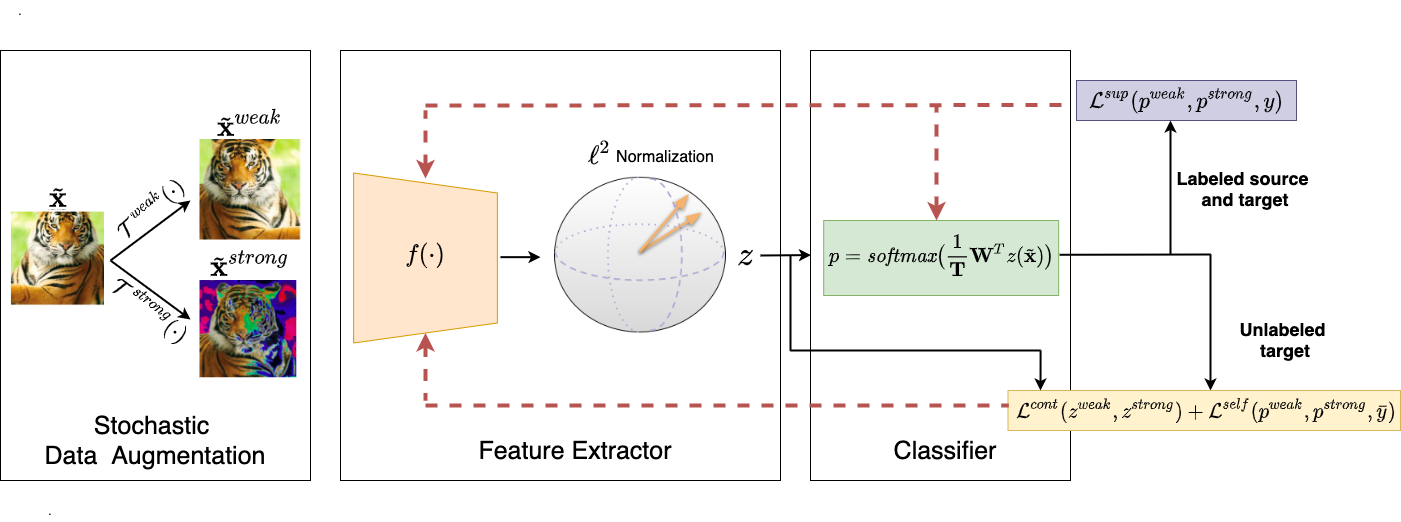}
  \caption{Overview of the Con$^{2}$DA framework. Our model takes weak and strong augmented samples and passes them through a feature extractor. The resulting vectors are normalized to the unit hypersphere. In the supervised case, the standard cross-entropy is applied for weak and strong augmented data. When labels are not available, a contrastive and a self-supervised loss functions are applied, leveraging both weak and strong augmented samples and extracting consistent feature representations in the normalized representation space. Notice that continuous black lines represent operations performed in the forward pass, while dashed red lines represent operations performed in the backward pass.}
  \label{fig:model}
   \vspace{-1.0em}

\end{figure*}

In a semi-supervised domain adaptation (SSDA) problem, we are given a source dataset composed by the images and their respective labels $\mathcal{D}^s = \{\mathbf{x}_{i}^s, y_{i}^s\}_{i=1}^{n^{s}}$, an unlabeled target dataset $\mathcal{D}^{u} = \{(\mathbf{x}_{i}^{u})\}^{n^u}_{i=1}$, and labeled target data $\mathcal{D}^{t} = \{(\mathbf{x}_{i}^{t}, y_{i}^{t})\}_{i=1}^{n^t}$. In SSDA, $n^{t}$ is assumed to be small (usually 1 or 3 labeled samples per class). For both domains we have the same $K$ classes, i.e. $y^s_{i}\in\{1,...,K\}$, $y^t_{i}\in\{1,...,K\}$. Our goal is to reduce the domain shift between $\mathcal{D}^{s}$ and $\{\mathcal{D}^{u}$, $\mathcal{D}^{t}\}$, to improve the performance of the model when evaluating on $\mathcal{D}^{u}$.

In this work we propose a method that learns \textbf{Con}sistent and \textbf{Con}trastive feature representations for SS\textbf{DA}. An overview of our method, named \textbf{Con$^{2}$DA} can be seen in Figure \ref{fig:model}. Our framework is composed of three major components:
\begin{itemize}
\item \textbf{Stochastic data augmentation:} Given an input $\mathbf{x} \in \{\mathcal{D}^{s}, \mathcal{D}^{u}, \mathcal{D}^{t}\}$, two random transformations  are applied, generating the two associated versions of the same sample $\tilde{\mathbf{x}}^{weak} \sim \mathcal{T}^{weak}(\mathbf{x})$ and $\tilde{\mathbf{x}}^{strong} \sim \mathcal{T}^{strong}(\mathbf{x})$. For $\mathcal{T}^{weak}(\cdot)$, we used simple transformations such as \textit{random cropping}, \textit{random horizontal flipping}, and \textit{Gaussian blur}. For $\mathcal{T}^{strong}(\cdot)$, we used the same augmentations, plus RandAugment \cite{cubuk_2020}. We show that by adopting an augmentation strategy that strongly perturbs input images, our algorithm improves in performance compared with the method using only weak augmentations (see Appendix D).
\item\textbf{Feature extractor:} We used a feature extractor $f(\cdot)$ that maps transformations $\tilde{\mathbf{x}} \in \{\tilde{\mathbf{x}}^{weak}, \tilde{\mathbf{x}}^{strong}\}$ into a representation vector $h = f(\tilde{\mathbf{x}}) \in \mathbb{R }^{d}$, where $d$ is the representation dimension (we set $d=256$ in all our experiments). This vector is normalized to the unit hypersphere using the $\ell _{2}$ normalization, that takes the form $z(\cdot) = \frac{f(\cdot)}{\lVert f(\cdot) \rVert}$. Depending on whether $\tilde{\mathbf{x}}$ is a labeled or unlabeled sample, different loss functions are optimized. 
\item \textbf{Classifier:} We pass the normalized feature representations through a linear classifier $\textbf{W} \in \mathbb{R}^{d\times K}$ and we scale the output by a temperature hyperparameter $T$. Using a softmax activation function we obtained the output probability distribution  $p(y|\tilde{\mathbf{x}}) = \textit{Softmax}\big(\frac{1}{\mathbf{T}}\mathbf{W}^{T}z(\tilde{\mathbf{x}})\big)$. It is worth noticing that for accurate classification, both the weight vector and the normalized features of the same class should be pointing to the same direction into the hypersphere. This way, each weight vector $\textbf{w}_{i} \in \textbf{W}$, $i \in \{1,...,K\}$ can be seen as a representative prototype of each class \cite{saito_2019}. 
\end{itemize}

\textbf{Supervised Loss Function:} For the labeled objects, we align associated data pairs towards their respective class prototype, regardless of whether the objects come from the source or target domain. Specifically, we sample a mini-batch of data  $\{(\textbf{x}_{i}, y_{i})\}^{N}_{i=1} \in \{\mathcal{D}^{s}, \mathcal{D}^{t}\}$, composed of $N/2$ number of source labeled objects and $N/2$ number of target labeled objects. For each sample $\mathbf{x}_{i}$, two stochastic augmentations are applied, obtaining a weak augmented version $\tilde{\mathbf{x}_{i}}^{weak}$, and a strong augmented version $\tilde{\mathbf{x}_{i}}^{strong}$. Both images are passed through the feature extractor and classifier, obtaining the probability vectors $p^{weak}_{i} = p(y_{i}|\tilde{\mathbf{x}}^{weak}_{i})$ and $p^{strong}_{i} = p(y_{i}|\tilde{\mathbf{x}}^{strong}_{i})$. Then, the loss function for each mini-batch can be computed using Equation \ref{eq:sup}:
\begin{equation} \label{eq:sup}
\mathcal{L}^{sup} = \frac{1}{N}\sum^{N}_{i=1} \big( \mathcal{L}_{ce}(p^{weak}_{i}, y_{i}) + \mathcal{L}_{ce}(p^{strong}_{i}, y_{i}) \big),
\end{equation}
where $\mathcal{L}_{ce}$ is a standard cross-entropy loss. Intuitively, by minimizing this objective with respect to the feature extractor and the linear classifier, we force the labeled feature representations and prototypes that represent the same class to point in the same direction into the hypersphere.

%For the unlabeled objects, we sample a mini-batch of size $N$ and the weak and strong transformation are applied to each sample $\mathbf{x}$, obtaining $\tilde{\mathbf{x}}^{weak}$ and $\tilde{\mathbf{x}}^{strong}$.
 \textbf{Unsupervised Loss Functions:} In the unsupervised case, we expect the distance of normalized feature representations of associated data pairs to be low, while we expect the distance with feature representations of objects that were generated from different inputs to be high. For this purpose, we sample a mini-batch $\{\textbf{x}_{i}\}^{N}_{i=1} \in \mathcal{D}^{u}$ of $N$ unlabeled target objects and we use the normalized temperature-scaled cross-entropy loss function (NT-Xent; \cite{chen_2020}), that takes two associated normalized feature representations $z^{weak}_{i} = z(\tilde{\textbf{x}}^{weak}_{i})$ and $z^{strong}_{i} = z(\tilde{\textbf{x}}^{strong}_{i})$,  and computes the cosine similarity between them $sim(z^{weak}_{i}, z^{strong}_{i}) = z^{weak}_{i} \cdot z^{strong}_{i}$. This similarity is forced to be high, while the similarities $sim(z_{i}, z_{a})$ between each feature representation $z_{i} \in \{z_{i}^{weak}, z_{i}^{strong}\}$ and the rest of the dissimilar augmented samples $z_{a} \in A(i) = \{z^{weak}_{1}, ..., z^{weak}_{i-1}, z^{weak}_{i+1}, ...,z^{weak}_{N}\} \cup \{z^{strong}_{1}, ..., z^{strong}_{i-1}, z^{strong}_{i+1}, ...,z^{strong}_{N}\}$  are forced to be low. The mathematical formulation for the NT-Xent loss can be written as follows:
\begin{equation} \label{eq:cont}
\mathcal{L}^{cont} = \frac{1}{N}\sum^{N}_{i=1} 
 -log{\frac{\exp(sim(z^{weak}_{i}, z^{strong}_{i})/T)}{\sum_{z_{a} \in A(i)}\exp(sim(z_{i},z_{a})/T)}},
\end{equation}
where $T$ denotes a temperature hyperparameter. Notice that this loss function is computed for both $z_{i}=z_{i}^{weak}$ and $z_{i}=z_{i}^{strong}$. 

Our model also computes an artificial label for each pair of strong and weak augmented unlabeled samples (i.e. pseudo-labeling \cite{mclachlan_1975, Lee_2013, sohn_2020}). This label can be obtained by computing the $\argmax$ function over the averaged model predictions for the $k$ classes, for a given pair of associated samples $\bar{y}_{i} = \argmax_{1\leq k \leq K} ((p^{weak}_{ik}+p^{strong}_{ik})/2)$. Similarly, an averaged model confidence can be computed as $\bar{p}_{i} = \max_{1\leq k \leq K} ((p^{weak}_{ik}+p^{strong}_{ik})/2)$. Then, for each mini-batch our method minimizes a self-supervised loss function with respect to the feature extractor:
\begin{equation} \label{eq:self}
\mathcal{L}^{self} =  \frac{1}{N} \sum^{N}_{i=1} \big(
\mathbbm{1}_{{\bar{p}_{i}\geq \tau}} 
\mathcal{L}_{ce}(p^{weak}_{i}, \bar{y}_{i}) +
\mathbbm{1}_{{\bar{p}_{i}\geq \tau}}
\mathcal{L}_{ce}(p^{strong}_{i}, \bar{y}_{i}) \big)
\end{equation}
where $\mathbbm{1}_{{\bar{p}_{i}\geq \tau}} \in \{0,1\}$ is the indicator function that takes value $1$ iff $\bar{p}_{i}$ is greater than  $\tau$, a probability threshold hyperparameter, and $0$ otherwise. For simplicity, we assume that $\bar{p}_{i}$ are valid \textit{one-hot} probability distributions for the cross-entropy loss.

The overall optimization procedure can be seen in Appendix E.

\section{Experiments} \label{sec:exp}
sAs in the standard SSDA scenario, we randomly chose one or three labeled training samples per target class for training (one and three-shot respectively). We also randomly selected three target labeled samples as the validation set. We used all the unlabeled data for training, and we reveal their labels to evaluate and report the final model performance.\\
\textbf{Datasets.} We performed experiments using three benchmark datasets: \textbf{DomainNet} \cite{peng_2019} a large-scale domain adaptation dataset that contains six domains and 345 classes in each domain. Following \cite{saito_2019}, four domains (R: Real, P: Painting, S: Sketch, C: Clipart), 126 classes, and 7 different adaptation scenarios are used for evaluation. \textbf{Office-Home} \cite{office-home} contains four domains (R: Real, A: Art, C: Clipart, P: Product) and 65 classes. We performed experiments using 12 adaptation scenarios. \textbf{Office31} \cite{saenko_2010} contains three domains (W: Webcam, D: DSLR, A: Amazon), and 31 classes. We used two adaptation scenarios for evaluation (D to A, and W to A).\\
\textbf{Baselines.} We compared our method against state-of-the-art SSDA and UDA approaches, as well as using no domain adaptation at all. The baselines for SSDA consist of the state-of-the-art methods MME \cite{saito_2018}, APE \cite{kim_2020}, and BiAT \cite{Jiang_2020}.  For UDA we compared with DANN \cite{ganin_2014}, ADR \cite{saito_2018}, and CDAN \cite{long_2018}. We trained UDA methods treating target labeled samples as if they were source labeled samples. Non adaptation methods consisted of models trained on all labeled samples using cross-entropy, leveraging only labeled data (S+T; \cite{yu_2019}), and considering unlabeled samples by minimizing the conditional entropy (ENT; \cite{grandvalet_2004}). \\
\textbf{Results}
A summary of our results is shown in Table \ref{table:results} (see Appendix C for complete result tables). For the DomainNet dataset, our model achieved competitive results compared to the rest of state-of-the-art methods for AlexNet and ResNet34 in both one and three shot cases. As can be seen for Office-Home our method outperforms the averaged results reported by previous state-of-the-art results by $1.3\%$ of accuracy in average. Finally, for Office31 our model outperforms previous state-of-the-art methods by a margin of $1.0\%$ and $1.4\%$ of accuracy in one-shot and three-shot scenarios respectively.

\begin{table}[h]
  \begin{center}
  \scalebox{0.75}{
  \begin{tabular}{|l||cc||cc||cc||cc|}
    \hline
    &\multicolumn{2}{c||}{ \textbf{DomainNet}}&
    \multicolumn{2}{c||}{ \textbf{Office-Home}}&
    \multicolumn{2}{c||}{ \textbf{Office31}} & 
    \multicolumn{2}{c|}{ \textbf{DomainNet}}\\ 

    Method     & 1-shot     & 3-shot & 1-shot & 3-shot &  1-shot     & 3-shot & 1-shot     & 3-shot \\ \hline
    & \multicolumn{6}{c||}{AlexNet} & \multicolumn{2}{c|}{ResNet34} \\ \hline

    S+T \cite{yu_2019} & 40.0 & 40.3  & 44.1 & 50.0 & 50.2 & 61.8 & 56.9 & 60.0  \\ 
    
    DANN \cite{ganin_2014} & 40.4 & 42.4  & 45.1 & 50.3 & 55.8 & 64.8 & 58.4 & 60.7\\ 

    ADR \cite{saito_2018} & 39.2 & 42.7  & 44.5 & 49.5 & 50.6 & 61.3 & 57.6 & 60.4 \\ 

    CDAN \cite{long_2018} & 39.1 & 41.0  & 41.2 & 46.2 & 49.5 & 60.9 & 62.5 & 66.5\\ 

    ENT \cite{grandvalet_2004} & 29.1 & 39.8  & 38.8 & 50.9 & 50.4 & 65.1 & 62.6 & 67.6  \\ 
    
    MME \cite{saito_2018} & 44.2 & 48.2  & 49.2 & 55.2 & 56.5 & 67.6 & 66.4 & 68.9  \\ 
    
    APE \cite{kim_2020} & 44.6 & 48.9 & - & 55.6 & - & 68.3 & 67.6 & \textbf{71.7}\\ 
    
    BiAT \cite{Jiang_2020} & \textbf{45.5} & 49.4 & 49.6 & \textbf{56.4} & 56.3 & 68.4 & 67.1 & 69.7 \\  \hline    

    Con$^{2}$DA (Ours) & \textbf{45.5} & \textbf{49.5} & \textbf{50.5} &
    55.8 &
    \textbf{57.3} & \textbf{69.8} & \textbf{68.4} & 71.4\\

    \hline    
%Sat. 19:13
  \end{tabular}}
    \caption{Mean Accuracy over the different adaptation settings for DomainNet, Office-Home and Office31. All experiments were performed using the same adaptations and data splits used by \cite{saito_2019, kim_2020}.}
  \label{table:results}
  \end{center}
  \vspace{-2.0em}
\end{table}

\section{Conclusions}
Despite the huge labeling efforts that deep neural network models need to achieve their best performances, SSDA methods can be applied to greatly reduce labeling costs. In this work, we present Con$^{2}$DA, a simple framework that learns from artificially augmented images using consistent and contrastive visual representation to improve generalization on target domain. We studied the main components of our model and we showed the effect of different combinations of augmentation and training strategies. Using these findings, we compared our methods with different SSDA and UDA methods and we obtained state-of-the-art performances in three commonly used domain adaptation benchmarks DomainNet, Office-Home, and Office31 by margins of $1.3\%$, $1.3\%$, and $1.4\%$ respectively.

\section{Acknowledgements}
We gratefully acknowledge support from ANID through the FONDECYT Initiation grant Nº11191130 and from the Data Science Unit at University of Concepción. G.C.V. acknowledges support from the ANID --Millennium Science Initiative Program--ICN12\_009 awarded to the Millennium Institute of Astrophysics (MAS).

\bibliographystyle{plain}
\bibliography{egbib}

\newpage
\section{Appendix}

\noindent \textbf{Appendix A: Implementation Details}: 

We implemented our model using PyTorch \cite{pytorch_2019}. For fair comparison with previous methods, we used AlexNet \cite{krizhevsky_2012} or ResNet34 \cite{he_2016} pretrained on Imagenet \cite{Russakovsky_2015} as the feature extractor. If the number of labeled objects class in the target domain is greater than 256, we set the batch size to 256. If not, we set the batch size to the number of target labeled objects. We sample two mini-batches, where the first mini-batch contains the same number of labeled objects per domain, and the second mini-batch is composed of only unlabeled target samples. We optimized our model using Adam \cite{kingma_2015} with hyperparameters $\beta_{1}=$ 0.9, $\beta_{2}=0.999$, and a learning rate of 0.00008 for all experiments. We decayed the learning rate with the cosine decay schedule without restarts as in \cite{loshchilov_2017}. In all scenarios, the augmentation policy was set to $H=1$ and the distortion magnitude to $M=10$ (details in Appendix A). The rest of the hyperparameters were set to ($T=0.05$, $\tau=0.9$) for DomainNet, ($T=0.3$, $\tau=0.95$) for Office-Home, and ($T=0.5$, $\tau=0.95$) for Office31. We did a forward pass using the two mini-batches (labeled source and target, and unlabeled target), calculated their respective losses, and did a backward pass. We called each of these forward-backward passes an iteration. We trained each model over 5000 iterations, setting a patience of 50 iterations for early stopping. \\

 \textbf{Appendix B: Hyperparameter Optimization}

We randomly sample three target labeled objects per class as the validation set and we use this set to set the hyperparameters temperature $T$ and threshold $\tau$. We selected the hyperparameters using different domains and adaptation scenarios. Specifically, for Office31 we used the W$\rightarrow$A scenario, and for Office-Home and DomainNet we used the R$\rightarrow$C scenario. We vary the hyperparameters from $T \in \{0.01, 0.05, 0.07, 0.1, 0.3, 0.5, 0.7, 0.9\}$ and $\tau \in \{0.8, 0.9, 0.95\}$. 

\begin{figure}
  \label{fig:hyperparams} 
  \begin{minipage}[b]{0.5\linewidth}
    \centering
    \includegraphics[width=1\linewidth]{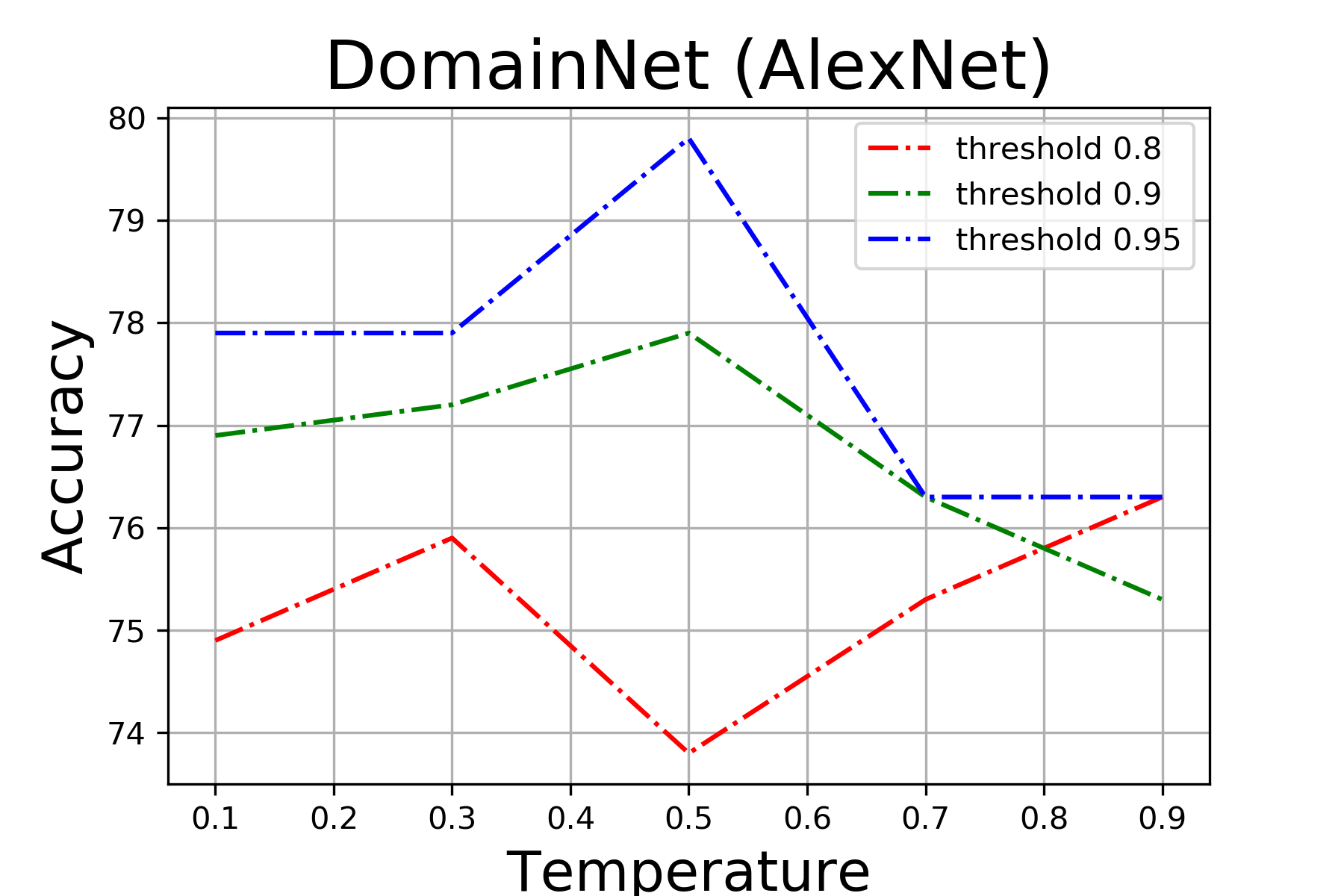} 
  \end{minipage}%%
  \begin{minipage}[b]{0.5\linewidth}
    \centering
    \includegraphics[width=1\linewidth]{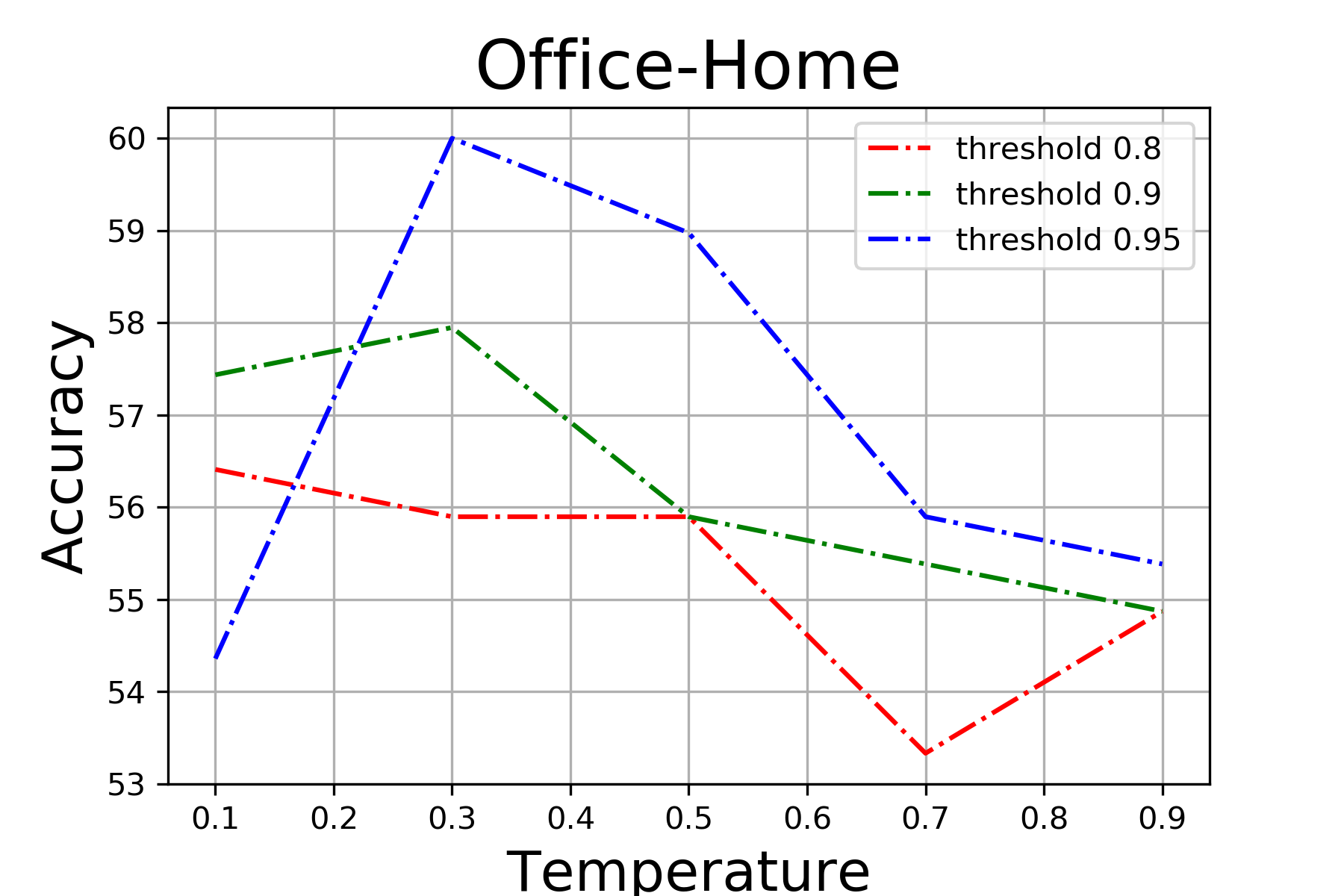} 
  \end{minipage} 
  \begin{minipage}[b]{0.5\linewidth}
    \centering
    \includegraphics[width=1\linewidth]{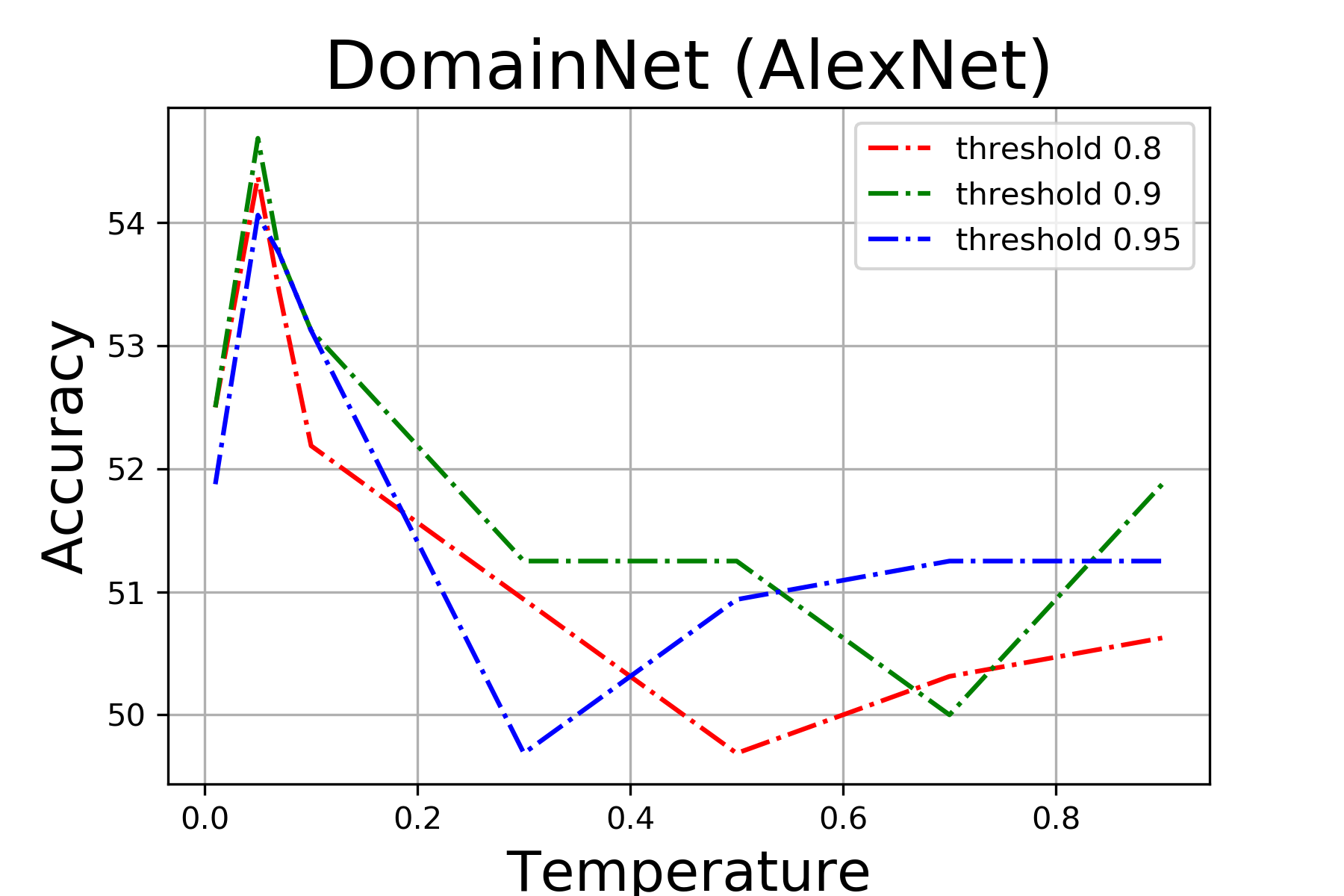} 
  \end{minipage}%% 
  \begin{minipage}[b]{0.5\linewidth}
    \centering
    \includegraphics[width=1\linewidth]{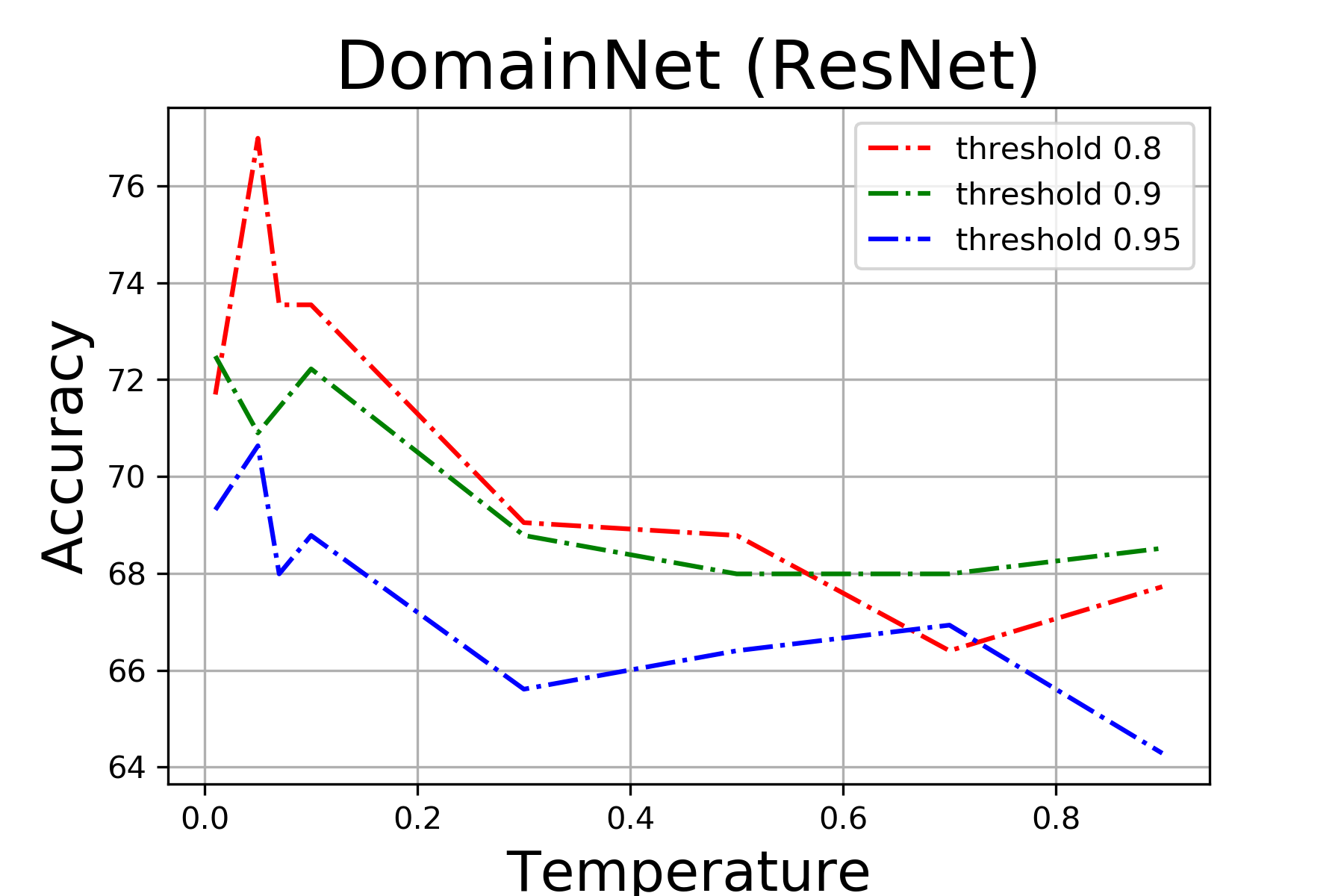} 
  \end{minipage} 
\caption{Accuracy for different values of the hyperparatemers threshold and temperature. Results computed for Office31, Office-Home and DomainNet (using AlexNet and ResNet34 model architectures).}
\end{figure}

Figure \ref{fig:hyperparams} shows the results in terms of accuracy each case. As can be seen, the best results on validation set are achieved using the hyperparameters ($T=0.05$, $\tau=0.9$) for DomainNet (AlexNet), ($T=0.05$, $\tau=0.8$) for DomainNet (ResNet34), ($T=0.3$, $\tau=0.95$) for Office-Home, and ($T=0.5$, $\tau=0.95$) for Office31. Therefore, we setted the hyperparameters to these values for the remaining adaptation scenarios in each case for the results.

\textbf{Appendix C: Results per adaptaton setting}

Results of our method on the DomainNet dataset are shown in Table \ref{table:multi}. In both the one and three-shot cases, our model achieved competitive results compared to the rest of state-of-the-art methods for AlexNet and ResNet34. As can be seen in Table \ref{table:office-home}, for Office-Home our method outperforms the averaged results reported by previous state-of-the-art results by margins between $0.1\%$ up to $1.3\%$ of accuracy. Finally, as can be seen in Table \ref{table:office}, for Office31 our model outperforms previous state-of-the-art methods by a margin of $1.0\%$ and $1.4\%$ of accuracy in one-shot and three-shot scenario respectively.

\begin{table*}[h]
  \begin{center}
  \scalebox{0.65}{
  \begin{tabular}{|l|cc|cc|cc|cc|cc|cc|cc||cc|}
    \hline
    &\multicolumn{2}{c|}{R$\rightarrow$ C}&
    \multicolumn{2}{c|}{R$\rightarrow$ P}&
    \multicolumn{2}{c|}{P$\rightarrow$ C}&
    \multicolumn{2}{c|}{C$\rightarrow$ S}&
    \multicolumn{2}{c|}{S$\rightarrow$ P}&
    \multicolumn{2}{c|}{R$\rightarrow$ S}&
    \multicolumn{2}{c||}{P$\rightarrow$ R}&
    \multicolumn{2}{c|}{Mean}\\

    Method     & 1-shot     & 3-shot & 1-shot     & 3-shot & 1-shot     & 3-shot & 1-shot     & 3-shot & 1-shot     & 3-shot & 1-shot     & 3-shot & 1-shot     & 3-shot & 1-shot     & 3-shot \\ \hline
    & \multicolumn{14}{c}{AlexNet} & & \\ \hline

    S+T \cite{yu_2019} & 43.3 & 47.1 & 42.4 & 45.0 & 40.1 & 44.9 & 33.6 & 36.4 & 35.7 & 38.4 & 29.1 & 33.3 & 55.8 & 58.7 & 40.0 & 43.4   \\ 
    
    DANN \cite{ganin_2014} & 43.3 & 46.1 & 41.6 & 43.8 & 39.1 & 41.0 & 35.9 & 36.5 & 36.9 & 38.9 & 32.5 & 33.4 & 53.6 & 57.3 & 40.4 & 42.4   \\ 

    ADR \cite{saito_2018} & 43.1 & 46.2 & 41.4 & 44.4 & 39.3 & 43.6 & 32.8 & 36.4 & 33.1 & 38.9 & 29.1 & 32.4 & 55.9 & 57.3 & 39.2 & 42.7   \\ 

    CDAN \cite{long_2018} & 46.3 & 46.8 & 45.7 & 45.0 & 38.3 & 42.3 & 27.5 & 30.2 & 33.7 & 33.7 & 28.8 & 31.3 & 56.7 & 58.7 & 49.1 & 41.0   \\ 

    ENT \cite{grandvalet_2004} & 37.0 & 45.5 & 35.6 & 42.6 & 26.8 & 40.4 & 18.9 & 31.1 & 15.1 & 29.6 & 18.0 & 29.6 & 52.2 & 60.0 & 29.1 & 39.8   \\ 
    
    MME \cite{saito_2018} & 48.9 & 55.6 & 48.0 & 49.0 & 46.7 & 51.7 & 36.3 & 39.4 & 39.4 & \textbf{43.0} & 33.3 & 37.9 & 56.8 & 60.7 & 44.2 & 48.2   \\ 
    
    APE \cite{kim_2020} & 47.7 & 54.6 & 49.0 & 50.0 & 46.9 & 52.1 & \textbf{38.5} & \textbf{42.6} & 38.5 & 42.2 & 33.8 & 38.7 & 57.5 & 61.4 & 44.6 & 48.9   \\ 
    
    BiAT \cite{Jiang_2020} & \textbf{54.2} & \textbf{58.6} & 49.2 & 50.6 & 44.0 & 52.0 & 37.7 & 41.9 & \textbf{39.6} & 42.1 & \textbf{37.2} & \textbf{42.0} & 56.9 & 58.8 & \textbf{45.5} & 49.4   \\  \hline    

    Con$^{2}$DA (Ours) & 50.0 & 54.5 & \textbf{50.3} & \textbf{51.6} & \textbf{47.7} & \textbf{53.5} & 36.9 & \textbf{42.6} & 35.4 & 40.3 & 35.9 & 39.2 & \textbf{62.7} & \textbf{64.8} & \textbf{45.5} & \textbf{49.5}  \\

    \hline
    \hline 
    & \multicolumn{14}{c}{ResNet34} & & \\ \hline
    
    S+T \cite{yu_2019} & 55.6 & 60.0 & 60.6 & 62.2 & 56.8 & 59.4 & 50.8 & 55.0 & 56.0 & 59.5 & 46.3 & 50.1 & 71.8 & 73.9 & 56.9 & 60.0   \\ 
    
    DANN \cite{ganin_2014} & 58.2 & 59.8 & 61.4 & 62.8 & 56.3 & 59.6 & 52.8 & 55.4 & 57.4 & 59.9 & 52.2 & 54.9 & 70.3 & 72.2 & 58.4 & 60.7   \\ 

    ADR \cite{saito_2018} & 57.1 & 60.7 & 61.3 & 61.9 & 57.0 & 60.7 & 51.0 & 54.4 & 56.0 & 59.9 & 49.0 & 51.1 & 72.0 & 74.2 & 57.6 & 60.4   \\ 

    CDAN \cite{long_2018} & 65.0 & 69.0 & 64.9 & 67.3 & 63.7 & 68.4 & 53.1 & 57.8 & 63.4 & 65.3 & 54.5 & 59.0 & 73.2 & 78.5 & 62.5 & 66.5   \\ 

    ENT \cite{grandvalet_2004} & 65.2 & 71.0 & 65.9 & 69.2 & 65.4 & 71.1 & 54.6 & 60.0 & 59.7 & 62.1 & 52.1 & 61.1 & 75.0 & 78.6 & 62.6 & 67.6   \\ 
    
    MME \cite{saito_2018} & 70.0 & 72.2 & 67.7 & 69.7 & 69.0 & 71.7 & 56.3 & 61.8 & \textbf{64.8} & 66.8 & 61.0 & 61.9 & 76.1 & 78.5 & 66.4 & 68.9   \\ 
    
    APE \cite{kim_2020} & 70.4 & \textbf{76.6} & 70.8 & \textbf{72.1} & \textbf{72.9} & \textbf{76.7} & 56.7 & 63.1 & 64.5 & 66.1 & 63.0 & \textbf{67.8} & 76.6 & \textbf{79.4} & 67.6 & \textbf{71.7}   \\ 
    
    BiAT \cite{Jiang_2020} & \textbf{73.0} & 74.9 & 68.0 & 68.8 & 71.6 & 74.6 & 57.9 & 61.5 & 63.9 & \textbf{67.5} & 58.5 & 62.1 & \textbf{77.0} & 78.6 & 67.1 & 69.7   \\  \hline    

    Con$^{2}$DA (Ours) & 71.3 & 74.2 & \textbf{71.8} & \textbf{72.1} & 71.1 & 75.0 & \textbf{60.0} & \textbf{65.7} & 63.5 & 67.1 & \textbf{65.2} & 67.1 & 75.7 & 78.6 & \textbf{68.4} & 71.4  \\
    
    \hline
%Sat. 19:13
  \end{tabular}}
    \caption{Accuracy on DomainNet ($\%$) using AlexNet \cite{krizhevsky_2012} and ResNet34 \cite{he_2016}.}
  \label{table:multi}
  \end{center}
  \vspace{-1.0em}
\end{table*}

\begin{table*}[ht]
  \begin{center}
  \scalebox{0.75}{
  \begin{tabular}{|c|c|c|c|c|c|c|c|c|c|c|c|c||c|}
    \hline
    \multicolumn{1}{|c|}{Method}&
    \multicolumn{1}{c|}{R$\rightarrow$ C}&
    \multicolumn{1}{c|}{R$\rightarrow$ P}&
    \multicolumn{1}{c|}{R$\rightarrow$ A}&
    \multicolumn{1}{c|}{P$\rightarrow$ R}&
    \multicolumn{1}{c|}{P$\rightarrow$ C}&
    \multicolumn{1}{c|}{P$\rightarrow$ A}&
    \multicolumn{1}{c|}{A$\rightarrow$ P}&
    \multicolumn{1}{c|}{A$\rightarrow$ C}&
    \multicolumn{1}{c|}{A$\rightarrow$ R}&
    \multicolumn{1}{c|}{C$\rightarrow$ R}&
    \multicolumn{1}{c|}{C$\rightarrow$ A}&
    \multicolumn{1}{c||}{C$\rightarrow$ P}&
    \multicolumn{1}{c|}{Mean}\\ 
    \hline
    & \multicolumn{12}{c}{One-shot} & \\ \hline

    S+T \cite{yu_2019} & 37.5 & 63.1 & 44.8 & 54.3 & 31.7 & 31.5 & 48.8 & 31.1 & 53.3 & 48.5 & 33.9 & 50.8 & 44.1 \\ 
    
    DANN \cite{ganin_2014} & 42.5 & 64.2 & 45.1 & 56.4 & 36.6 & 32.7 & 43.5 & 34.4 & 51.9 & 51.0 & 33.8 & 49.4 & 45.1   \\ 

    ADR \cite{saito_2018} & 37.8 & 63.5 & 45.4 & 53.5 & 32.5 & 32.2 & 49.5 & 31.8 & 53.4 & 49.7 & 34.2 & 50.4 & 44.5 \\ 

    CDAN \cite{long_2018} & 36.1 & 62.3 & 42.2 & 52.7 & 28.0 & 27.8 & 48.7 & 28.0 & 51.3 & 41.0 & 26.8 & 49.9 & 41.2  \\ 

    ENT \cite{grandvalet_2004} & 26.8 & 65.8 & 45.8 & 56.3 & 23.5 & 21.9 & 47.4 & 22.1 & 53.4 & 30.8 & 18.1 & 53.6 & 38.8  \\ 
    
    MME \cite{saito_2018} & 42.0 & 69.6 & \textbf{48.3} & 58.7 & 37.8 & 34.9 & 52.5 & 36.4 & 57.0 & \textbf{54.1} & \textbf{39.5} & 59.1 & 49.2    \\ 

    BiAT \cite{Jiang_2020} & - & - & - & - & - & - & - & - & - & - & - & - & 49.6   \\ \hline

     Con$^{2}$DA (Ours) & \textbf{43.7} &  \textbf{70.7} &  47.8 & \textbf{60.4} & \textbf{39.8} & \textbf{36.8} & \textbf{62.5} & \textbf{36.9} & \textbf{58.5} & 53.0 & 35.4 & \textbf{60.0} & \textbf{50.5} \\

    \hline
    \hline

     & \multicolumn{12}{c}{Three-shot} & \\ \hline

    S+T \cite{yu_2019} & 44.6 & 66.7 & 47.7 & 57.8 & 44.4 & 36.1 & 57.6 & 38.8 & 57.0 & 54.3 & 37.5 & 57.9 & 50.0   \\ 
    
    DANN \cite{ganin_2014} & 47.2 & 66.7 & 46.6 & 58.1 & 44.4 & 36.1 & 57.2 & 39.8 & 56.6 & 54.3 & 38.6 & 57.9 & 50.3   \\ 

    ADR \cite{saito_2018} & 45.0 & 66.2 & 46.9 & 57.3 & 38.9 & 36.3 & 57.5 & 40.0 & 57.8 & 53.4 & 37.3 & 57.7 & 49.5 \\ 

    CDAN \cite{long_2018} & 41.8 & 69.9 & 43.2 & 53.6 & 35.8 & 32.0 & 56.3 & 34.5 & 53.5 & 49.3 & 27.9 & 56.2 & 46.2  \\ 

    ENT \cite{grandvalet_2004} & 44.9 & 70.4 & 47.1 & 60.3 & 41.2 & 34.6 & 60.7 & 37.8 & 60.5 & 58.0 & 31.8 & 63.4 & 50.9   \\ 
    
    MME \cite{saito_2019} & 51.2 & 73.0 & 50.3 & 61.6 & 47.2 & 40.7 & 63.9 & 43.8 & 61.4 & \textbf{59.9} & \textbf{44.7} & 64.7 & 55.2    \\ 
    
    APE \cite{kim_2020} & 51.9 & \textbf{74.6} & \textbf{51.2} & 61.6 & 47.9 & \textbf{42.1}& 65.5 & 44.5 & 60.9 & 58.1 & 44.3 & 64.8 & 55.6   \\ 
    
    BiAT \cite{Jiang_2020} & - & - & - & - & - & - & - & - & - & - & - & - & \textbf{56.4}   \\ \hline

    Con$^{2}$DA (Ours) & \textbf{52.3} & 73.5 &  49.1 & \textbf{64.4} & \textbf{49.3} & 38.2 & \textbf{66.4} & \textbf{47.7} & \textbf{62.4} & \textbf{59.9} & 39.9 & \textbf{66.1}&  55.8\\

    \hline    
%Sat. 19:13
  \end{tabular}}
    \caption{Accuracy on Office-Home ($\%$) 3-shot using AlexNet \cite{krizhevsky_2012}.}
  \label{table:office-home}
  \end{center}
  \vspace{-1.0em}
\end{table*}

\begin{table}[h]
  \begin{center}
  \scalebox{0.75}{
  \begin{tabular}{|l|cc|cc||cc|}
    \hline
    &\multicolumn{2}{c|}{W$\rightarrow$ A}&
    \multicolumn{2}{c||}{D$\rightarrow$ A}&
    \multicolumn{2}{c|}{Mean}\\ 

    Method     & 1-shot     & 3-shot & 1-shot     & 3-shot & 1-shot     & 3-shot \\ \hline

    S+T \cite{yu_2019} & 50.4 & 61.2 & 50.0 & 62.4 & 50.2 & 61.8  \\ 
    
    DANN \cite{ganin_2014} & 57.0 & 64.4 & 54.5 & 65.2 & 55.8 & 64.8\\ 

    ADR \cite{saito_2018} & 50.2 & 61.2 & 50.9 & 61.4 & 50.6 & 61.3 \\ 

    CDAN \cite{long_2018} & 50.4 & 60.3 & 48.5 & 61.4 & 49.5 & 60.9 \\ 

    ENT \cite{grandvalet_2004} & 50.7 & 64.0 & 50.0 & 66.2 & 50.4 & 65.1  \\ 
    
    MME \cite{saito_2018} & 57.2 & 67.3 & 55.8 & 67.8 & 56.5 & 67.6  \\ 
    
    APE \cite{kim_2020} & - & 67.6 & - & 69.0 & - & 68.3  \\ 
    
    BiAT \cite{Jiang_2020} & 57.9 & 68.2 & 54.6 & 68.5 & 56.3 & 68.4 \\  \hline    

    Con$^{2}$DA (Ours) & \textbf{58.3} & \textbf{69.8} & \textbf{56.2} & \textbf{69.7} & \textbf{57.3} & \textbf{69.8}\\

    \hline    
%Sat. 19:13
  \end{tabular}}
    \caption{Accuracy on Office-31 ($\%$) using AlexNet.}
  \label{table:office}
  \end{center}
  \vspace{-2.0em}
\end{table}

\textbf{Appendix D: Ablation Study}

\begin{figure}[h]
\centering
  \includegraphics[width=0.8\linewidth]{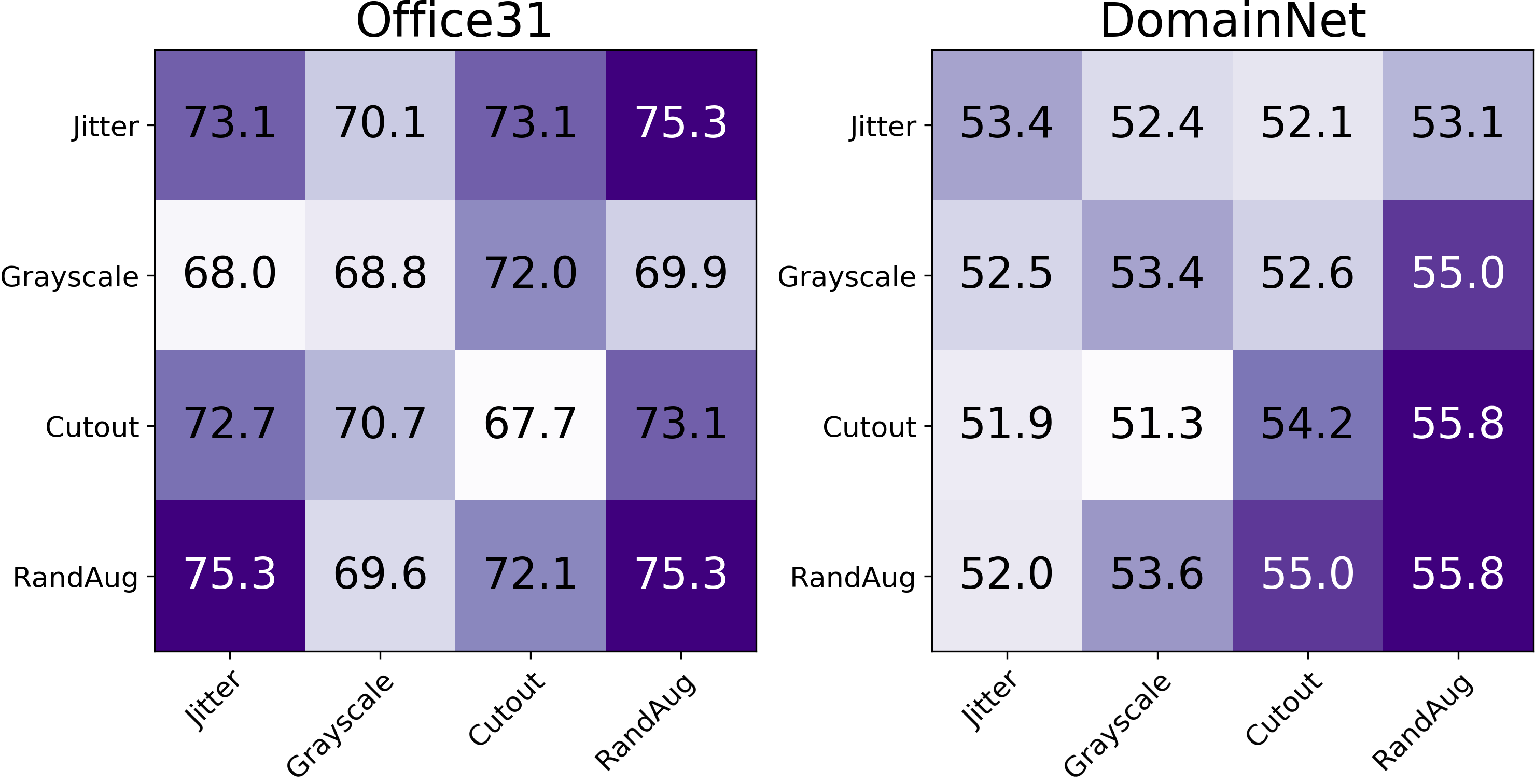}
  \caption{Comparison of different augmentation policies for Office31 (W$\rightarrow$ A) and DomainNet (R$\rightarrow$ C) in a three-shot scenario. Performance measured on the validation set.}
  \label{fig:augmentations}
   \vspace{-1.0em}

\end{figure}

In order to understand the key components of our method, we performed an ablation study to compare gains in performance using variations of our algorithm. We first investigated the performance of different strong augmentation policies. We started by first measuring the performance on a baseline where no strong augmentations are applied (i.e both $\tilde{\mathbf{x}}^{weak}$ and $\tilde{\mathbf{x}}^{strong}$ are samples from $\mathcal{T}^{weak}$). In a three-shot scenario, the accuracy on validation set is $72.0\%$ for Office31 in the W$\rightarrow$A setting, and $50.5\%$ for DomainNet in the R$\rightarrow$ C setting. We added strong augmentation policies to $\tilde{\mathbf{x}}^{strong}$ such as color jitter, random grayscale, Cutout \cite{devries_2017}, and RandAugment \cite{cubuk_2020}. The performance between combinations of these augmentation policies is shown in Figure \ref{fig:augmentations}. As can be seen, our model consistently improved the baseline and the rest of the augmentation strategies in both settings by using RandAugment as a strong augmentation policy. Using RandAugment, we also investigate the model performance removing components from the loss function. Firstly, we removed the contrastive loss function defined in Equation \ref{eq:cont} (w/o $\mathcal{L}_{cont}$), the self-supervised loss function defined in Equation \ref{eq:self}, w/o $\mathcal{L}_{self}$. Then, we forced the prototypes to be unit vectors by normalizing the linear classifier $\textbf{W}$, and we computed the cosine similarity between the normalized linear classifier and feature vector (w cosine). Results of these variations are shown in Table \ref{table:ablation}. As can be seen, our algorithm is benefited from the use of contrastive and self-supervised loss. Finally, the linear classifier without normalization helped our model to obtain significant performance improvements. 

\begin{table}[ht]
  \begin{center}
  \scalebox{0.75}{
  \begin{tabular}{|l|c|c|}
    \hline
    &\multicolumn{1}{c|}{W$\rightarrow$ A}&
    \multicolumn{1}{c|}{R$\rightarrow$ C} \\  \hline   
    
   w/o $\mathcal{L}_{cont}$ & 66.7 & 54.5 \\ 
   
   w/o $\mathcal{L}_{self}$ & 73.1 & 51.8\\
   
   w cosine & 72.0 & 55.7 \\ \hline 

    Con$^{2}$DA & \textbf{75.3} & \textbf{55.8} \\

    \hline    
%Sat. 19:13
  \end{tabular}}
    \caption{Accuracy ($\%$) of different training strategies for Office31 (W$\rightarrow$ A) and DomainNet (R$\rightarrow$ C) in a three-shot scenario. Performance measured on the validation set.}
  \label{table:ablation}
  \end{center}
  \vspace{-2.0em}
\end{table}

\textbf{Appendix E: Algorithm} 
Algorithm 1 summarizes the Con$^{2}$DA training procedure.

\begin{algorithm} \label{algorithm}
    \caption{Con$^{2}$DA Training procedure.}
  \begin{algorithmic}[1]
    \INPUT Mini-batches $\{\textbf{x}_{i}, y_{i}\}^{N}_{i=1} \in \{\mathcal{D}^{s}, \mathcal{D}^{t}\}$ and $\{\textbf{x}^{u}_{i}\}^{N}_{i=1} \in \mathcal{D}^{u}$. Hyperparameters $\tau$ and $T$. Transformations $\mathcal{T}^{weak}$ and $\mathcal{T}^{strong}$. Feature extractor $f$ and linear classifier $\mathbf{W}$. 
    \WHILE{not converged}:
      \FOR{$i \in \{1,...,N\}$}
      \STATE{\textcolor{mygray}{\# Augmentations for labeled data}}
      \STATE{$\tilde{\textbf{x}}^{weak}_{i} \sim \mathcal{T}^{weak}(\textbf{x}_{i})$}
      \STATE{$\tilde{\textbf{x}}^{strong}_{i} \sim \mathcal{T}^{strong}(\textbf{x}_{i})$}
      \STATE{\textcolor{mygray}{\# Probability vectors for labeled samples.}}
      \STATE{$p^{weak}_{i} = p(y_{i}| \tilde{\textbf{x}}^{weak}_{i})$}
      \STATE{$p^{strong}_{i} = p(y_{i}| \tilde{\textbf{x}}^{strong}_{i})$}
      \STATE{\textcolor{mygray}{\# Augmentations for unlabeled data}}
      \STATE{$\tilde{\textbf{x}}^{u ,weak}_{i} \sim \mathcal{T}^{weak}(\textbf{x}_{i}^{u})$}
      \STATE{$\tilde{\textbf{x}}^{u ,strong}_{i} \sim \mathcal{T}^{strong}(\textbf{x}_{i}^{u})$}
      \STATE{\textcolor{mygray}{\# Normalized feature representations.}}
      \STATE{$z^{u,weak}_{i} = f(\tilde{\textbf{x}}^{u,weak}_{i}))/ \lVert f(\tilde{\textbf{x}}^{u,weak}_{i}))\rVert$}
       \STATE{$z^{u,strong}_{i} = f(\tilde{\textbf{x}}^{u,strong}_{i})) / \lVert f(\tilde{\textbf{x}}^{u,strong}_{i}))\rVert$}
      \STATE{\textcolor{mygray}{\# Probability vectors for unlabeled samples.}}
      \STATE{$p^{u,weak}_{i} = p(y_{i}| \tilde{\textbf{x}}^{u,weak}_{i})$}
      \STATE{$p^{u,strong}_{i} = p(y_{i}| \tilde{\textbf{x}}^{u,strong}_{i})$}
      \STATE{\textcolor{mygray}{\# Averaged prediction and pseudo-label.}}
      \STATE{$\bar{p}_{i} = \max_{1\leq k \leq K} ((p^{u,weak}_{ik}+p^{u,strong}_{ik})/2)$}
      \STATE{$\bar{y}_{i} = \argmax_{1\leq k \leq K} ((p^{u,weak}_{ik}+p^{u,strong}_{ik})/2)$}
    \ENDFOR{}
   \FOR{$i \in \{1,...,N\}$}
   \STATE{$sim_{ii} = z^{u,weak}_{i} \cdot z^{u,strong}_{i}$}
   \STATE{Define $A(i)$}
   \FOR{$z_{a} \in A(i)$}
   \STATE{$sim_{ia} = z^{u}_{i} \cdot z^{a}$}
   \ENDFOR{}
%   \STATE{$sim(z^{u}_{i}, z^{u}_{j}) = z^{u}_{i} \cdot z^{u}_{j} $}
    \ENDFOR{}
    \STATE{Compute Equations \ref{eq:sup}, \ref{eq:cont}, and \ref{eq:self}}
    \STATE{Update parameters of $f$ and $W$ to minimize Eq. \ref{eq:sup}.}
    \STATE{Update parameters of $f$ to minimize Eq. \ref{eq:cont} and \ref{eq:self}. }
    \ENDWHILE 
    \OUTPUT{Trained feature extractor $f$ and classifier $\mathbf{W}$}
  \end{algorithmic}
\end{algorithm}

\end{document}